\pdfoutput=1
\documentclass[letterpaper]{article} 
\usepackage{aaai20}  
\usepackage{times}  
\usepackage{helvet} 
\usepackage{courier}  
\usepackage[hyphens]{url}  
\usepackage{graphicx} 
\usepackage{graphicx}  
\usepackage{graphicx}
\usepackage{amsmath}
\usepackage{xcolor}
\usepackage{cool}
\usepackage{tabularx}
\usepackage{multirow}
\usepackage{array}
\usepackage{booktabs}
\usepackage{makecell}
\usepackage{mathrsfs}
\usepackage[ruled]{algorithm2e}
\usepackage{amsmath}
\usepackage{subcaption}

\DeclareMathOperator*{\argmin}{arg\,min}
\newcommand{\citet}[1]{\citeauthor{#1} \shortcite{#1}}
\newcommand{\citep}{\cite}

\newcommand{\ignore}[1]{}
\title{Smart Predict-and-Optimize for Hard Combinatorial
Optimization Problems}
\author{Jayanta Mandi,\textsuperscript{\rm 1} Emir Demirovi\'{c},\textsuperscript{\rm 2} Peter. J Stuckey,\textsuperscript{\rm 2} Tias Guns\textsuperscript{\rm 1}\\ \textsuperscript{\rm 1} Data Analytics Laboratory, Vrije Universiteit Brussel \{jayanta.mandi,tias.guns\}@vub.be\\
\textsuperscript{\rm 2} University of Melbourne \{emir.demirovic,pstuckey\}@unimelb.edu.au
}
 
\begin{document}
\maketitle
\begin{abstract}
Combinatorial optimization assumes that all parameters of the optimization problem, e.g. the weights in the objective function, are fixed. Often, these weights are mere estimates and increasingly machine learning techniques are used to for their estimation.
Recently, Smart Predict and Optimize (SPO) has been proposed for problems with a linear objective function over the predictions, more specifically linear programming problems. It takes the regret of the predictions on the linear problem into account, by repeatedly solving it during learning.
We investigate the use of SPO to solve more realistic discrete optimization problems. The main challenge is the repeated solving of the optimization problem. To this end, we investigate ways to relax the problem as well as warm-starting the learning and the solving.
Our results show that even for discrete problems it often suffices to train by solving the relaxation in the SPO loss. Furthermore, this approach outperforms the state-of-the-art approach of \citeauthor{wilder2018melding}.
We experiment with weighted knapsack problems as well as complex scheduling problems, and show for the first time that a predict-and-optimize approach can successfully be used on large-scale combinatorial optimization problems.
\end{abstract}

\section{Introduction}
Combinatorial optimization aims to optimize an objective function over a set of feasible solutions defined on a discrete space. Numerous real-life decision-making problems can be formulated as combinatorial optimization problems~\cite{korte2012combinatorial,trevisan2011combinatorial}.
In the last decade, development of time-efficient algorithms for combinatorial optimization problems paved the way for these algorithms to be widely utilized in industry, including, but not limited to,
in resource allocation~\cite{angalakudati2014business}, efficient energy scheduling~\cite{ngueveu2016scheduling}, price optimization~\cite{ferreira2015analytics}, sales promotion planning~\cite{cohen2017impact}, etc.

The last decade has, in parallel, witnessed a tremendous growth in machine learning (ML) methods, which can produce very accurate predictions by leveraging historical and contextual data.
In real-world applications, not all parameters of an optimization problem are known at the time of execution and predictive ML models can be used for estimation of those parameters from historical data. For instance, ~\citeauthor{cohen2017impact} first predicted future demand of products using an ML model and then use the predicted demand to compute the optimal promotion pricing scheme over the products through non-linear integer programming. 

When predictive ML is followed by optimization, it is often assumed that improvements in the quality of the predictions (with respect to some suitable evaluation metric) will result in better optimization outcomes. 
%
However, ML models make errors and 
the impact of prediction errors is not uniform throughout the underlying solution space, for example, overestimating the highest-valued prediction might not change a maximization problem outcome, while underestimating it can. 
Hence, a better prediction model may not ensure a better outcome in the optimization stage. In this regard, \citet{ifrim2012properties} experienced that a better predictive model does not always translate to optimized energy-saving schedules. 


The alternative is to take the effect of the errors on the optimization outcome into account during learning. 
In the context of linear programming problems, \citeauthor{elmachtoub2017smart}
proposed an approach, called ``Smart Predict and Optimize'' (SPO), for
training ML models by minimizing a convex surrogate loss function which
considers the outcome of the optimization stage. Specifically they consider
optimization problems where 
predictions occur as weights that are linear in the objective.

In this work, we build on that approach and consider discrete combinatorial optimization problems. 
Indeed, the SPO loss is valid for any optimization problem with a linear
objective over predictions, 
where the constraints implicitly define a convex region. 
Furthermore, any black box optimization method can be used as only its outcome is used to compute the (sub)gradient of the loss.

The main challenge is the computational cost of the repeated solving of the optimization problem during training, namely once for every evaluation of the loss function on an instance. For NP-hard problems, this may quickly become infeasible.
In order to scale up to large problem instances, we investigate the importance of finding the optimal discrete solution for learning, showing that continuous relaxations are highly informative for learning. 
Furthermore, we investigate how to speed up the learning by transfer learning from easier-to-learn models as well as method for speeding up the solving by warmstarting from earlier solutions. Our approach outperforms the state-of-the-art Melding approach~\cite{wilder2018melding} in most cases, and for the first time we are able to show the applicability of predict-and-optimize on large scale combinatorial instances, namely from the ICON energy-cost aware scheduling challenge~\cite{csplib:prob059}. 

\section{Related Work}

Predict-and-optimize problems arise in many applications. Current practice is to use a two-stage approach where the ML models are trained independent of the optimization problem. As a consequence, the ML models do not account for the optimization tasks~\cite{wang2006cope,mukhopadhyay2017prioritized}. 
In recent years there is a growing interest in decision-focused learning~\cite{elmachtoub2017smart,demirovicinvestigation,wilder2018melding}, that aims to couple ML and decision making.

In the context of portfolio optimization, \citet{bengio1997using} report a deep learning model fails to improve future profit when trained with respect to a standard ML loss function, but a profit-driven loss function turns out to be effective. 
\citet{kao2009directed} consider an unconstrained quadratic optimisation problem, where the predicted values appear linearly with the objective. They train a linear model with respect to a combination of  prediction error and optimization loss. They do not mention how this can be applied to optimization problems with constraints.

A number of works aim to exploit the fact that the KKT conditions of a quadratic program (QP) define a system of linear equations around the optimal points. For instance,
\citet{donti2017task} propose a framework which computes the gradient of the solution of the QP with respect to the predictions by applying the implicit function theorem to differentiate through the KKT conditions around the optimal point. \citet{wilder2018melding} use the same approach, and propose its use for linear programs by adding a small quadratic term to convert it into a concave QP. They also propose a specialisation of it for submodular functions. 

Our work builds on the SPO approach of \citet{elmachtoub2017smart}, where the authors provide a framework to train an ML model, which learns with respect to the error in the optimization problem. This is investigated for linear optimization problems with a convex feasible region. 
We will use the approach for discrete combinatorial problems with a linear objective. They are computationally expensive to solve, e.g. often $\mathcal{NP}$-hard.
The decision variables and search space of these problems is discrete, meaning gradients can not be computed in a straightforward manner. However, the SPO approach remains applicable as we will see.

\citet{demirovicinvestigation} investigate the prediction+optimisation
problem for the knapsack problem, and prove that optimizing over predictions
are as valid as stochastic optimisation over learned distributions, in case
the predictions are used as weights in a linear objective. They further
investigate possible learning approaches, and classified them into three
groups: \textit{indirect} approaches, which do not use knowledge of the
optimisation problem; \textit{semi-direct} approaches, which encode knowledge of the
optimisation problem, such as the importance of ranking and \textit{direct} approaches which encode or use the optimisation problem in the learning in some way~\cite{demirovicinvestigation}. Our approach is a \textit{direct} approach and we examine how to combine the best of such techniques in order to scale to large and hard combinatorial problems.


\section{Problem Formulation and Approach}

\subsection{Optimizing a parameterized problem}

Traditional optimization algorithms work under the assumption that all the parameters are known precisely. But in a predict-and-optimize setup we assume some parameters of the optimization problem are not known. 

We formalize a combinatorial optimization problem as follows : \begin{equation}
v^*(\theta) \equiv \argmin_v f(v,\theta)\ s.t.\  C(v,\theta) \label{oracle}
\end{equation}  where
$\theta$ defines the set of parameters (coefficients) of the optimization problem,
$v$ are the decision variables,
$f(v,\theta)$ is an objective to be minimized, and
$C(v,\theta)$ is a (set of) constraints that determine(s) the feasible region; hence $v^*()$ is an oracle that returns the optimal solution.

Consider, the $0$-$1$ knapsack, where a set of items, with their values and wights, are provided. The objective is to select a subset of items respecting a capacity constraint on the sum of weights  so that the total value of the subset is maximized. The parameter set $\theta$ of the problem consists of the value and weight of each item and the total capacity. The decision variable set consists of $0$-$1$ decision variable for each item, $f$ is a linear sum of the variables and the item values and $C$ describing the capacity constraint.

We decompose $\theta = \theta_s\cup \theta_u$ where $\theta_s$ are the set of parameters that are observed (e.g. the weights and capacity of a knapsack) and $\theta_u$ are the set of unobserved parameters (the value of the knapsack items).

To predict the unobserved parameters $\theta_u$, some attributes correlated with them are observed. We are equipped with a training dataset $ D: \{(x_1,\theta_{u_1}),...,(x_n,\theta_{u_n})\}$ where $x_i$'s are vectors of attributes correlated to $\theta_{u_i}$. An ML model $m$ is trained to generate a prediction $\hat{\theta}_u = m(x;\omega) $. The model $m$ is characterized by a set of learning parameters $\omega$.
E.g. in linear regression the $\omega$ encompasses the slope and the intercept of the regression line. Once the predicted $\hat{\theta}_u$ are obtained, $\theta = \theta_s\cup \hat{\theta}_u$ is used for the optimization problem.  

To ease notation, we will write $\hat{\theta}=(\theta_s, \hat{\theta}_u)$, containing both the observed parameters and predicted parameters. Recall from Equation~\eqref{oracle} that $v^*(\hat{\theta})$ is the optimal solution using parameters $\hat{\theta}$. Then, the objective value of this solution is $f(v^*(\hat{\theta}),\theta)$. Whereas if the actual $\theta$ is known \emph{a priori},
one could obtain the actual optimal solution $v^*({\theta})$. The difference in using the predicted instead of the actual values is hence measured by
\begin{align}
regret(\theta,\hat{\theta}) \equiv f(v^*(\hat{\theta}),\theta) - f(v^*(\theta),\theta) \label{eq:regret}
\end{align}
Ideally the aim of the training should be to generate predictions which minimize this regret on unseen data.

\subsection{Two Stage Learning}
First we formalize a two-stage approach, which is widely used in industry  and where the prediction and the optimization are performed in a decoupled manner. First the predictive model is trained with the objective of minimizing the expected loss for a suitable choice of loss function $ \mathcal{L}(\theta_{u},\hat{\theta}_u)$
 
For regression, if we use squared-error as the loss function, model parameters $\omega$ are estimated by minimizing  the  Mean Squared Error (MSE) on the training data:
\begin{equation}
\mathcal{L}_{MSE} = \frac{1}{n} \sum_{i=1}^n \frac{(\theta_{u_i} - \hat{\theta}_{u_i})^2}{2}
\end{equation}

%

\subsubsection{Training by gradient descent}

\begin{algorithm}
\SetAlgoLined
\SetKwRepeat{Repeat}{repeat }{until convergence} 
\Repeat{}
{Sample $N$ training datapoints

\For {$i$ in $1,...,N$}
{
predict $\hat{\theta}_{u_i} $ using current $\omega$
$\nabla\mathcal{L}_i \leftarrow   (\hat{\theta}_{u_i} - \theta_{u_i})$ //gradient of $\mathcal{L}_{MSE}$
}
$\nabla\mathcal{L}=\frac{ \sum_{i=1}^N \nabla\mathcal{L}_i }{N}$ 
$ \omega \leftarrow \omega -\alpha* \nabla\mathcal{L} * \frac{\partial \hat{\theta}_u  }{\partial \omega}$ 
}
 \caption{Stochastic Batch gradient descent for the two-stage learning for regression tasks (batchsize:$N$) and learning rate $\alpha$}
 \label{Algo:BGD}
\end{algorithm}

The process of estimating ${\omega}$ to minimize the loss function is  executed through stochastic gradient descent algorithms\footnote{There are different variations of sgd, for a detailed discussion refer \cite{ruder2016overview}},where at each epoch, ${\omega}$ are updated after calculating the gradient $\nabla \mathcal{L}$  of the loss function with respect to the predictions as shown in Algorithm~\ref{Algo:BGD}.

\ignore{
{\color{red} [ }There are several variants of gradient descent algorithm and mini-batch gradient descent is the most widely used to train ML algorithms. In mini-batch gradient descent, each update is performed after calculating the loss function and its gradient over a small number of datapoints (typically 32 to 256). There are many extensions~\cite{ruder2016overview} of standard gradient-descent which have become popular due to their success in
solving various ML problems.{\color{red} ] }
}

The advantage of the two-stage approach is that training by gradient descent is straightforward. In the prediction stage the objective is to minimize the MSE-loss to generate accurate predictions without considering their impact on the final solution. Clearly this 
 does \emph{not} require solving the optimization problem during training.

\subsubsection{Model validation and early stopping with regret}
It is common practice to perform model validation on a separate validation set while training the model. 
{\it Early stopping}~\cite{bishop2006pattern} is the practice of choosing the epoch with the smallest loss on the validation set as the final output of the training procedure. The objective is to avoid overfitting on training data.
The performance on the validation set is also used to select hyperparameters of the ML models~\cite{bergstra2012random}.

Considering the final task, in our setup, is minimizing the regret, we modify the two stage learning by measuring regret on the validation set for early stopping and hyperparameter selection. We call this the \textbf{MSE-r} approach.
It is more computationally expensive than MSE given that computing regret on the validation data for every epoch requires solving the optimization problems each time.

\subsection{Smart Predict then Optimize (SPO)}
\label{SPO}

\begin{figure}
  \centering \includegraphics[width=0.6\columnwidth]{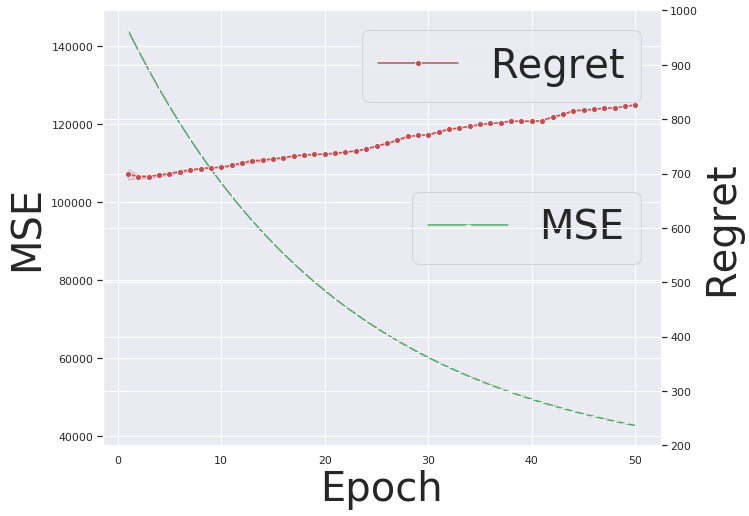}
  \caption{\small MSE (left axis) versus Regret (right axis) while training a knapsack instance; with no correlation and worsening regret.} 
  \label{RegretvsMSE}
\end{figure}

The major drawback of the two-stage approach is it does not aim to minimize the regret, but minimizes the error between $\theta_u$ and $\hat{ \theta}_u$ directly. As Figure~\ref{RegretvsMSE} shows, minimizing loss between $\theta_u$ and $\hat{ \theta}_u$ does not necessarily result in minimization of the regret. Early stopping with regret can avoid worsening results, but can not improve the learning. The SPO framework proposed by \citeauthor{elmachtoub2017smart} addresses this by integrating prediction and optimization. 

Note, to minimize $regret(\theta,\hat{\theta})$ directly we have to find the gradient of it with respect to $\hat{\theta}$ which requires differentiating the \emph{argmin} operator $v^*(\hat{\theta})$ in Eq.~\ref{oracle}. This differentiation may not be feasible as $v^*(\hat{\theta})$ can be discontinuous in $\hat{\theta}$ and exponential in size. Consequently we can not train an ML model to minimize the regret through gradient descent.

The SPO framework integrates the optimization problem into the loop of gradient descent algorithms in a clever way. In their work, \citeauthor{elmachtoub2017smart} consider an optimization problem with a convex feasible region $S$ and a linear objective :
\begin{equation}
v^* (\theta) \equiv \argmin_{v \in S} \theta^\top v
\end{equation} 
where the cost vector $\theta$ is not known beforehand.
Following Eq.~\eqref{eq:regret}, the regret for such a problem when using predicted values  $\hat{\theta}$ instead of actual $\theta$ is:  $\theta^\top (v^*(\hat{\theta}) -  v^*(\theta))$, which as discussed is not differentiable. 
To make it differentiable, they use a convex surrogate upper bound of the regret function, which they name the SPO+ loss function $\mathcal{L}_{SPO+}(\theta,\hat{\theta})$. The gradient of $\mathcal{L}_{SPO+}(\theta,\hat{\theta})$ may not exist as it also involves the  \emph{argmin} operator. However, they have shown that $v^*(\theta) - v^*(2\hat{\theta}-\theta)$ is a subgradient of $\mathcal{L}_{SPO+}(\theta,\hat{\theta})$, 
that is 
\begin{equation}\label{eq:subgradient}
g(\theta,\hat{\theta}) = v^*(\theta) - v^*(2\hat{\theta}-\theta), \quad g(\theta,\hat{\theta}) \in \nabla { \mathcal{L}_{SPO+}}(\theta,\hat{\theta})
\end{equation}

The subgadient formulation is the key to bring the optimization problem into the loop of gradient descent as shown in algorithm~\ref{Algo:SPO}.
\begin{algorithm}
\SetAlgoLined
\SetKwRepeat{Repeat}{repeat }{until convergence} 
\Repeat{}
{Sample $N$ training datapoints

\For {$i$ in $1,...,N$}
{
predict $\hat{\theta}_{u_i} $ using current $\omega$

compute $v^*(2\hat{\theta}-\theta)$
$\nabla\mathcal{L}_i \leftarrow   v^*(\theta) - v^*(2\hat{\theta}-\theta)$ //sub-gradient
} 
$\nabla\mathcal{L}=\frac{ \sum_{i=1}^N \nabla\mathcal{L}_i }{N}$

$ \omega \leftarrow \omega -\alpha* \nabla\mathcal{L} * \frac{\partial \hat{\theta}_u  }{\partial \omega}$; 
}
 \caption{Stochastic Batch gradient descent for the SPO approach for regression tasks (batchsize:$N$) and learning rate $\alpha$ }
 \label{Algo:SPO}
\end{algorithm}
The difference between algorithm~\ref{Algo:BGD} and algorithm~\ref{Algo:SPO} is in their (sub)gradients. In Algorithm~\ref{Algo:BGD}, the MSE gradient is the signed difference between the actual values and predicted ones; in Algorithm~\ref{Algo:SPO} the SPO subgradient is the difference of an optimization solution obtained using the actual parameter values and another solution obtained using a convex combination of the predicted values and the true values.

For the $0$-$1$ knapsack problem, the solution of a knapsack instance is a $0$-$1$ vector of length equal to the size of the set, where $1 $ represents the corresponding item is selected. In this case, the subgradient is the element-wise difference between the two solutions and if the solution using the transformed predicted values is the same as the solution using actual values, all entries of the subgradient are zero. In essence, the non-zero entries in the subgradient indicate places where the two solutions contradict.

Note, to compute the subgradient for this SPO approach, the optimization problem $v^*(2\hat{\theta}-\theta)$ needs to be solved for each training instance, while $v^*(\theta)$ can be precomputed and cached. Moreover, one training instance typically contains multiple predictions. 
For example, if we consider a 0-1 knapsack problem with 10 items, then one training instance always contains 10 value predictions, one for each item. Furthermore, the dataset may contain thousands of training instances of 10 values each. Hence, one iteration over the training data (one \textit{epoch}) requires solving hundreds of knapsack problems.
As an ML model is trained over several epochs, clearly the training
process is computationally expensive.


%

\subsection{Combinatorial problems and scaling up}
We observe that the SPO approach and its corresponding loss function places no restriction on the type of oracle $v^*()$ used. Given that our target task is to minimize the regret of the combinatorial problem, an oracle that solves the combinatorial optimisation problem is the most natural choice. We call this approach \textit{SPO-full}.

\paragraph{Weaker oracles}
Repeatedly solving combinatorial problems is computationally expensive. Hence for large and hard problems, it is necessary to look at ways to reduce the solving time.

As there is no restriction on the oracle used, we consider using weaker oracles during training.  NP-hard problems that have a polynomial (bounded) approximation algorithm could use the approximation algorithm in the loss as a proxy instead. For example, in case of knapsack, the \textit{greedy} algorithm~\cite{dantzig}. For mixed integer programming (MIP) formulations, a natural weaker oracle to use is the continuous relaxation of the problem. While disregarding the discrete part, relaxations can often identify what part of the problem is trivial (variable assignments close to 0 or 1) from what part is non-trivial. For example for knapsack, the continuous relaxation leads to very similar solutions compared to the greedy algorithm. Note that we always use the same oracle for $v^*(\theta)$ and $v^*(2\hat{\theta}-\theta)$ when computing the loss. We call the approach of using the continuous relaxation as oracle $v^*()$ \textit{SPO-relax}.

In case of weak MIP relaxations, one can also use a cutting plane algorithm in the root node and use the resulting tighter relaxation thereof~\cite{mipaal}.
Other weaker oracles could also be used, for example setting a time-limit on an any-time solver and using the best solution found, or a node-limit on search algorithms. In case of mixed integer programming, we can also set a gap tolerance, which means the solver does not have to prove optimality. We call this \textit{SPO-gap}. 
For stability of the learning, it is recommended that the solution returned by the oracle does not vary much when called with (near) identical input.

Apart from changing what is being solved, we also investigate ways to warmstart the learning, and to warmstart across solver calls:

\paragraph{Warmstarting the learning}
We consider warmstarting the learning by transfer learning~\cite{pratt1996survey}, that is, to train the model with an easy to compute loss function, and then continue training it with a more difficult one. In our case, we can pre-train the model using MSE as loss, which means the predictions will already be more informed when we start using an SPO loss afterwards.

More elaborate learning schemes are possible, such as curriculum learning~\cite{thrun2012learning,pratt1996survey} where we gradually move from easier to harder to compute loss functions, e.g. by moving from \textit{SPO-relax} to \textit{SPO-gap} for decreasing gaps to \textit{SPO-full}. As we will see later, this is not needed for the cases we studied.

\paragraph{Warmstarting the solving}
When computing the loss, we must solve both $v^*(\theta)$ using the true values $\theta$, and $v^*(2\hat{\theta}-\theta)$. Furthermore, we know that an optimal solution to $v^*(\theta)$ is also a valid (but potentially suboptimal) solution to $v^*(2\hat{\theta}-\theta)$ as only the coefficients of the objective differ. Furthermore, if this is an optimal solution to the latter than we would achieve $0$-regret, hence we can expect the solution of $v^*(\theta)$ to be of decent quality for $v^*(2\hat{\theta}-\theta)$ too.

We hence want to aide the solving of $v^*(2\hat{\theta}-\theta)$ by using the optimal solution of $v^*(\theta)$. One way to do this for CP solvers is solution-guided search~\cite{cp2018b}. For MIP/LP we can use warmstarting~\cite{yildirim2002warm,zeilinger2011real}, that is, to use the previous solution as starting point for MIP. In case of linear programming (and hence the relaxation), we can reuse the basis of the solution.

An alternative is to use the true solution to compute a bound on the objective function. Indeed, as the solution to $S = v^*(\theta)$ is valid for $v^*(2\hat{\theta}-\theta)$ and has an objective value of $f(S, (2\hat{\theta}-\theta))$. Hence, we can use this as a bound on the objective and potentially cut away a large part of the search space.

While the true solutions $v^*(\theta)$ can be cached, we must compute this solution once for each training instance $(x,\theta)_i$, which may already take significant time for large problems. We observe that only the objective changes between calls to the oracle $v^*$, and hence any previously computed solution is also a candidate solution for the other calls. We can hence use warmstarting for any solver call after the first, and from any previously computed solution so far.

\section{Experimental Evaluation}


We consider three types of combinatorial problems: unweighted and weighted knapsack and energy-cost aware scheduling. Below we briefly discuss the problem formulations:

\paragraph{Unweighted/weighted knapsack problem}
The knapsack problem can be formalized as $argmax_X V^\top X \text{ s.t. } W^\top X \leq c$. The values $V$ will be predicted from data and weights $W$ and capacity $c$ are given. In the \textit{unweighted knapsack}, all weights $W$ are 1 and the problem is polynomial time solvable. \textit{Weighted knapsacks} are NP-hard and it is known that the computational difficulty increases with the correlation between weights and values~\cite{pisinger2005hard}. We generated mildly correlated knapsacks as follows: for each of the 48 half-hour slots we assign a weight $w_i$ by sampling from the set $\{3,5,7\} $, then we multiply each profit value $v_i$ by its corresponding weight and include some randomness by adding Gaussian noise $ \xi \sim \mathbb{N}(0,25)$ to each $v_i$ before multiplying by weight.

In the unweighted case, 
we consider 9 different capacity values $c$, namely from 5 to 45 increasing by 5. 
For the weighted knapsack experiment, 
we consider 7 different capacity values from 30 to 210 increasing by 30.

\paragraph{Energy-cost aware scheduling}
It is a resource-constrained job scheduling problem where the goal is to minimize (predicted) energy cost, it is described in the CSPLib as problem 059~\cite{csplib:prob059}. In summary, we are given a number of machines and have to schedule a given number of tasks, where each task has a duration, an earliest start and a latest end, resource requirement and a power usage. Each machine has a resource capacity constraint. We omit startup/shutdown costs. No task is allowed to stretch over midnight between two days and cannot be interrupted once started, nor migrate to another machine. Time is discretized in $t$ timeslots and a schedule has to be made over all timeslots at once.
For each timeslot, a (predicted) energy cost is given and the objective is to minimize the total energy cost of running the tasks on the machines. 

We consider two variants of the problem: \textit{easy} instances consisting of 30 minute timeslots (e.g. 48 timeslots per day), and \textit{hard} instances as used in the ICON energy challenge consisting of 5 minute timeslots (288 timeslots per day). The easy instances have 3 machines and respectively 10, 15 and 20 tasks. The hard instances each have 10 machines and 200 tasks.


\paragraph{Data} Our data is drawn from the Irish Single Electricity Market Operator (SEMO) \citep{ifrim2012properties}.
This dataset consists of historical energy price data at 30-minute intervals starting from Midnight 1st November, 2011 to 31st December, 2013.
Each instance of the data has calendar attributes; day-ahead estimates of weather characteristics; SEMO day-ahead forecasted energy-load, wind-energy production and prices; and actual wind-speed, temperature, $CO_2$ intensity and price. Of the actual attributes, we keep only the actual price, and use it as a target for prediction. For the hard scheduling instances, each 5-minute timeslots have the same price as the 30-minute timeslot it belongs to, following the ICON challenge.


\subsubsection*{Experimental setup}
For all our experiments, we use a linear model without any hidden layer as the underlying predictive model. Note, the SPO approach is a model-free approach and it is compatible wih any deep neural network; but earlier work \cite{ifrim2012properties} showed accuracy in predictions is not effective for the downstream optimization task. 
For the experiments, we divide our data into three sets: training (70\%), validation (10\%) and test (20\%), and evaluate the performance by measuring regret on the test set. Training, validation and test data covers 552, 60 and 177 days of energy data respectively. 

Our model is trained by batch gradient descent, where each batch corresponds to one day, that is, $48$ consecutive training instances namely one for each half hour of that day. This batch together forms one set of parameters of one optimisation instance, e.g. knapsack values or schedule costs. 
The learning rate and momentum for each model are selected through a grid search based on the regret on the validation dataset. The best combination of parameters is then used in the experiments shown.

Solving the knapsack instances takes sub-second time per optimisation instance, solving the easy scheduling problems takes 0.1 to 2 seconds per optimisation instance and for the hard scheduling problems solving just the relaxation already costs 30 to 150 seconds per optimisation instance. For the latter, this means that merely \textit{evaluating} regret on the test-set of 177 optimisation instances takes about 3 hours. With 552 training instances, one epoch of training requires 9 hours.

For all experiments except the hard instances, we repeat it 10 times and report on the mean and standard deviation.

We use the \emph{Gurobi} optimization-solver for solving the combinatorial problems, the \emph{nn module in Pytorch} to implement the predictive models and the \emph{optim module in Pytorch} for training the models with corresponding loss functions. Experiments were run on Intel(R) Xeon(R) CPU E3-1225 v5 @ 3.30GHz processors with 32GB  memory \footnote{The code of our
experiments is available at \url{https://github.com/JayMan91/aaai_predit_then_optimize.git}}.

\begin{figure}
     \centering
 \begin{subfigure}[t!]{0.3\columnwidth}
         \centering
         \includegraphics[width=\columnwidth]{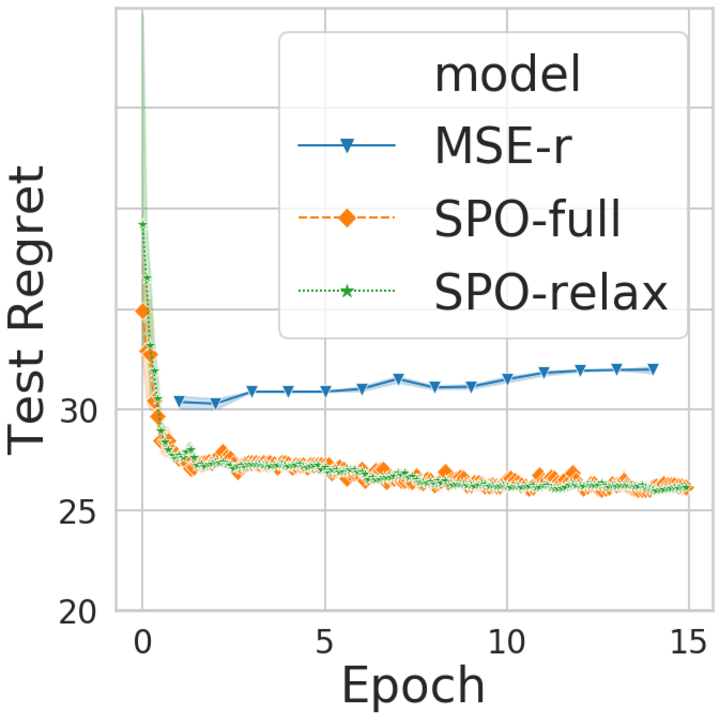}
         \caption{Unweighted \\knapsack(cap:10)}
         \label{fig:unweighted_SPOvsSPOrelax_learrningcurve}
     \end{subfigure} %
     ~
     \begin{subfigure}[t!]{0.3\columnwidth}
         \centering
         \includegraphics[width=\columnwidth]{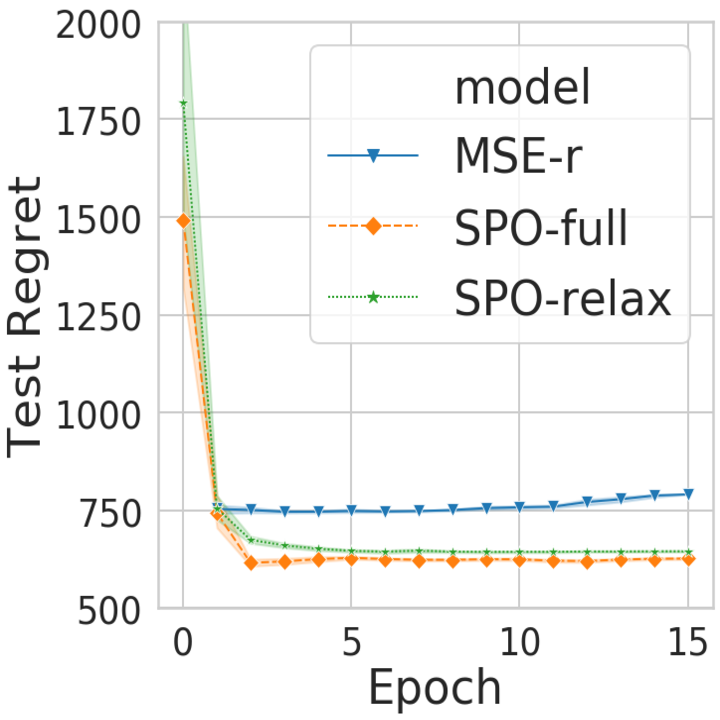}
         \caption{Weighted \\Knapsack (cap:60)}
         \label{fig:weighted_SPOvsSPOrelax_learrningcurve}
     \end{subfigure}%
    ~
     \begin{subfigure}[t!]{0.3\columnwidth}
         \centering
         \includegraphics[width=\columnwidth]{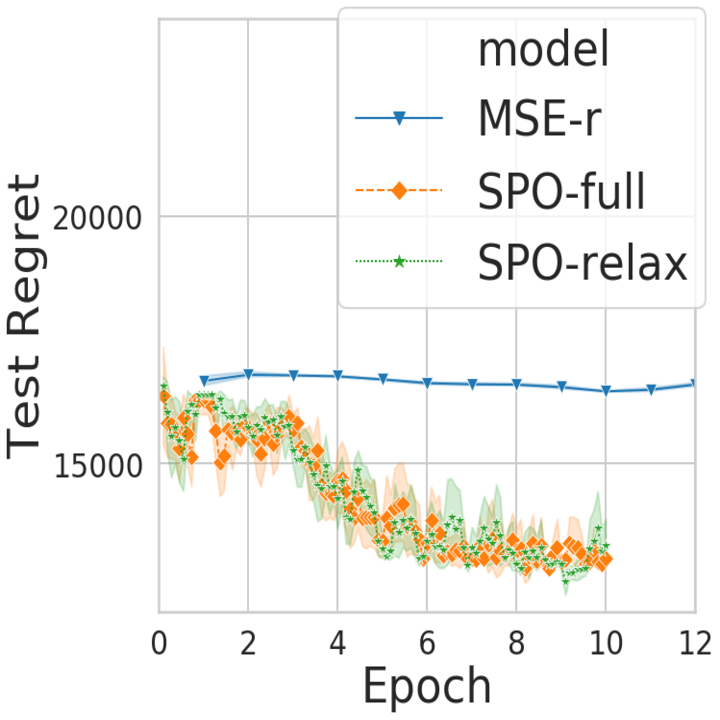}
         \caption{EnergySchedule \\(1st instance)}
         \label{fig:Energy_SPOvsSPOrelax_learrningcurve}
     \end{subfigure}
     \caption{Per epoch learning curves of MSE-r, SPO-full and SPO-relax}
        
        \label{fig:SPOvsSPOrelax_learrningcurve}
\end{figure}

\begin{figure}
     \centering
 \begin{subfigure}[t!]{0.3\columnwidth}
         \centering
         \includegraphics[width=\columnwidth]{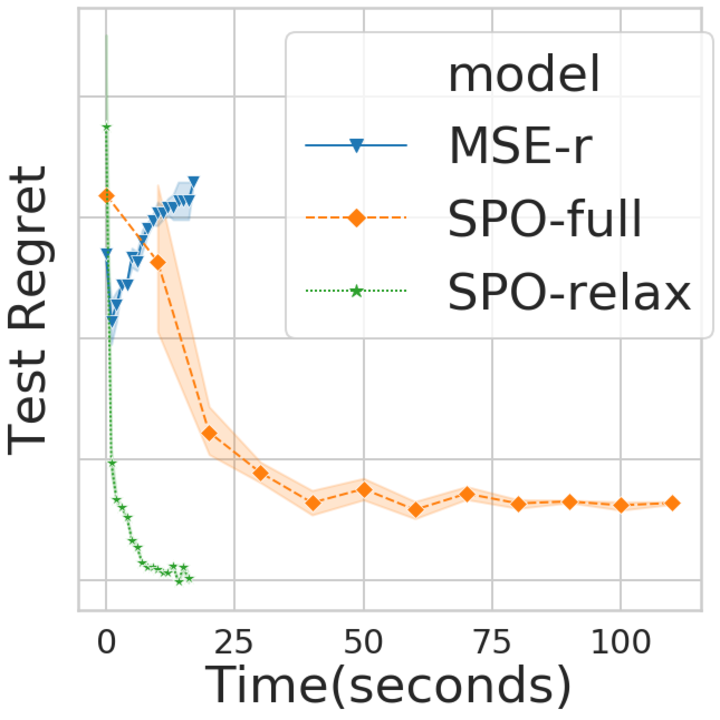}
         \caption{Unweighted\\knapsack(cap:10)}
         \label{fig:unweighted_SPOvsSPOrelax_time}
     \end{subfigure} %
     ~
     \begin{subfigure}[t!]{0.3\columnwidth}
         \centering
         \includegraphics[width=\columnwidth]{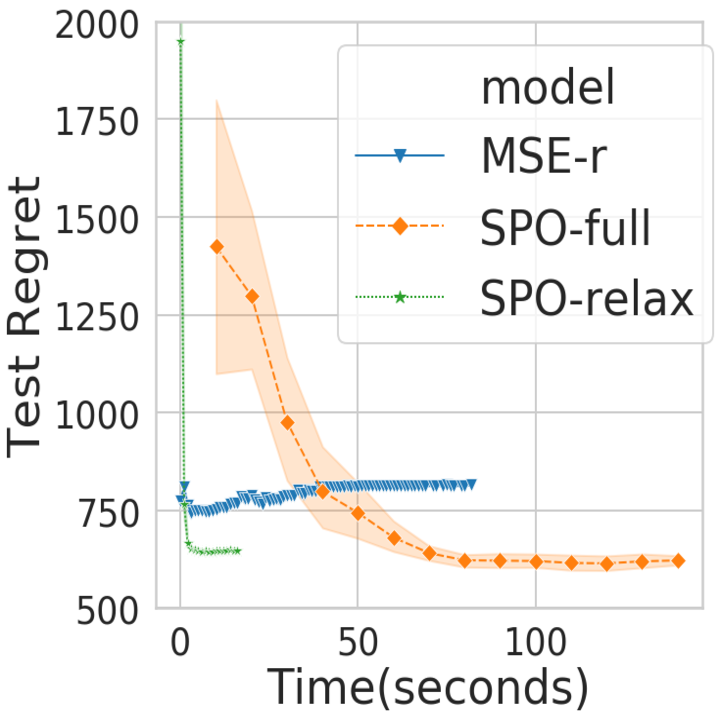}
         \caption{Weighted \\Knapsack(cap:60)}
          \label{fig:weighted_SPOvsSPOrelax_time}
     \end{subfigure}%
    ~
     \begin{subfigure}[t!]{0.3\columnwidth}
         \centering
         \includegraphics[width=\columnwidth]{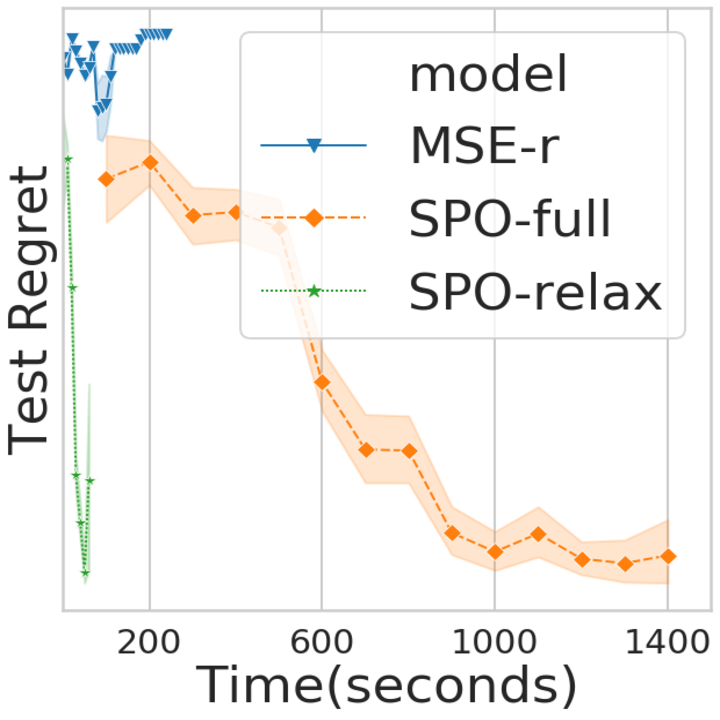}
         \caption{EnergySchedule \\(1st instance)}  \label{fig:Energy_SPOvsSPOrelax_time}
     \end{subfigure}
        \caption{Per second learning curves of MSE-r, SPO-full and SPO-relax}
        \label{fig:SPOvsSPOrelax_time}
\end{figure}


\subsection{RQ1: exact versus weaker oracles} 
The first research question is what the loss in accuracy of solving the full discrete problems (SPO-full) versus solving only the relaxation (SPO-relax) during training is. Together with this, we look at what the gain is in terms of reduction-in-regret over time. We visualise both through the learning curves as evaluated on the test set. For all methods, we compute the exact regret on the test-set, e.g. by fully solving the instances with the predictions.

Figure~\ref{fig:SPOvsSPOrelax_learrningcurve} shows the learning curves over epochs, where one epoch is one iteration over all instances of the training data, for three problem instances. We also include the regret of MSE-r as baseline. In all three case we see that MSE-r is worse, and stagnates or slightly decreases in performance over the epochs. This validates the use of MSE-r where the best epoch is chosen retrospectively based on a validation set. 
It also validates that the SPO-surrogate loss function captures the essence of the intended loss, namely regret.

We also see, surprisingly, that SPO-relax achieves very similar performance to SPO-full on all problem instances in our experiments. This means that even though SPO can reason over exact discrete solutions, reasoning over these continuous relaxation solutions is sufficient for the loss function to guide the learning to where it matters.


The real advantage of SPO-relax over SPO-full is evident from Figure~\ref{fig:SPOvsSPOrelax_time}. Here we present the test regret not against the number of epochs
but against the model run-time. MSE-r here includes the time to compute the regret on the validation set as needed to choose the best epoch. These figures show SPO-relax runs, and hence converges, much quicker in time than SPO. This is thanks to the fact that solving the continuous relaxation can be done in polynomial time while the discrete problem is worst-case exponential. SPO-relax is slower than MSE-r but the difference in quality clearly justifies it.

In the subsequent experiments, we will use SPO-relax only.

\subsection{RQ2 benefits of warmstarting}
\begin{table}
\begin{tabular}{crrrr}
\toprule
Instance & Baseline & MSE-warmstart & \makecell{Warmstart \\ from earlier basis}\\
\midrule 
1 & 6.5 (1.5) sec & 8 (0.5) sec & 1.5 (0.2) sec \\
\midrule
2 & 7 (1.5) sec & 6 (1.0) sec & 1 (0.2) sec \\
\midrule
3 & 10 (0.5) sec & 12 (1.0) sec & 2.5 (0.1) sec\\
\bottomrule
\end{tabular} 
\caption{Comparison of per epoch average (sd) runtime of warmstart strategies}
 \label{tab:Warmstart}
\end{table}
As \textit{baseline} we use the standard SPO-relax approach. We test \textit{warmstarting the learning} by first training the network with MSE as loss function for 6 epochs, 
after which we continue learning with SPO-relax. We indicate this approach by \textit{MSE-warmstart}.
We summarizes the effect of warmstarting in Table~\ref{tab:Warmstart},
We observe that warmstarting from MSE results in a slightly faster start in the initial seconds, but this has no benefit, nor penalty over the longer run. Warmstarting from an earlier basis, after the MIP pre-solving, did result in runtime improvements overall.

We also attempted warmstarting by adding objective cuts, but this slowed down the solving of the relaxation, often doubling it, because more iterations were needed. 

\subsection{RQ3: SPO versus QPTL} 
Next, we compare our approach against the state-of-the-art QP approach (QPTL) of \citet{wilder2018melding} which proposes to transform the discrete linear integer program into a continuous QP by taking the continuous relaxation and a squared L$2$-norm of the decision variables $||x||_2^2$. This makes the problem quadratic and twice differentiable allowing them to use the differntiable QP solver~\cite{donti2017task}.

\begin{figure}
     \centering
 \begin{subfigure}[t!]{0.3\columnwidth}
         \centering
         \includegraphics[width=\columnwidth]{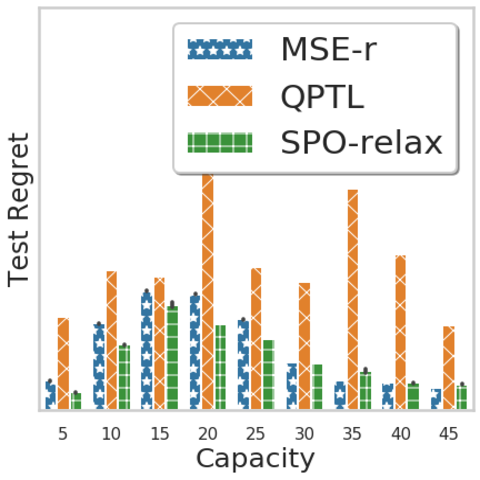}
         \caption{Unweighted knapsack}
         \label{fig:unweighted_SPOvsQPTL}
     \end{subfigure} %
     ~
     \begin{subfigure}[t!]{0.3\columnwidth}
         \centering
         \includegraphics[width=\columnwidth]{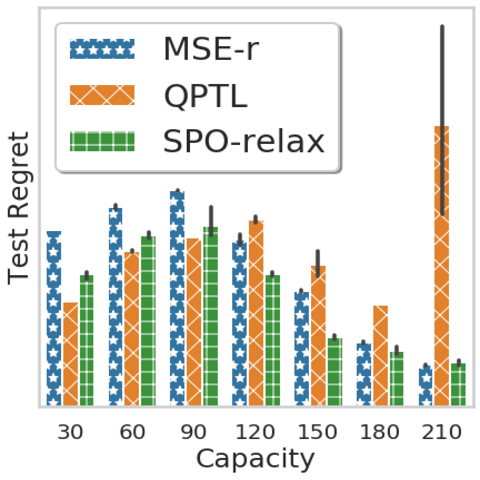}
         \caption{Weighted Knapsack}
         \label{fig:weighted_SPOvsQPTL}
     \end{subfigure}%
    ~
     \begin{subfigure}[t!]{0.3\columnwidth}
         \centering
         \includegraphics[width=\columnwidth]{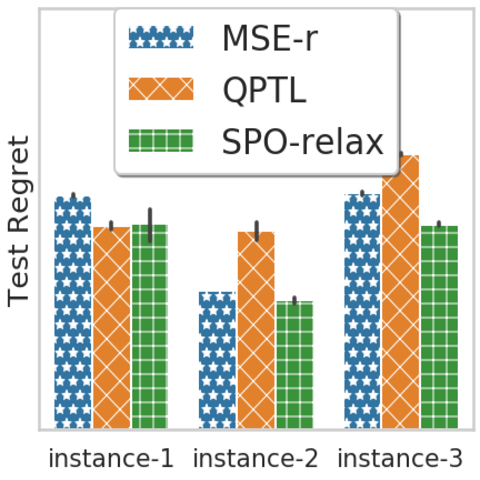}
         \caption{Energy Scheduling}
         \label{fig:Energy_SPOvsQPTL}
     \end{subfigure}
        \caption{MSE-r, SPO-relax and QPTL, all instances}
        \label{fig:SPOvsQPTL}
\end{figure}

Figure~\ref{fig:SPOvsQPTL} shows the average regret on all unweighted and weighted knapsack instances and easy scheduling instances. We can see that for unweighted knapsack, SPO-relax almost always outperforms the other methods, while QPTL performs worse than MSE-r. For weighted knapsacks, SPO-relax is best for all but the lower capacities. For these lower capacities, QPTL is better though its results worsen for higher capacities. 

\begin{figure}
     \centering
     \begin{subfigure}[t!]{0.14\columnwidth}
         \centering
         \includegraphics[width=\columnwidth]{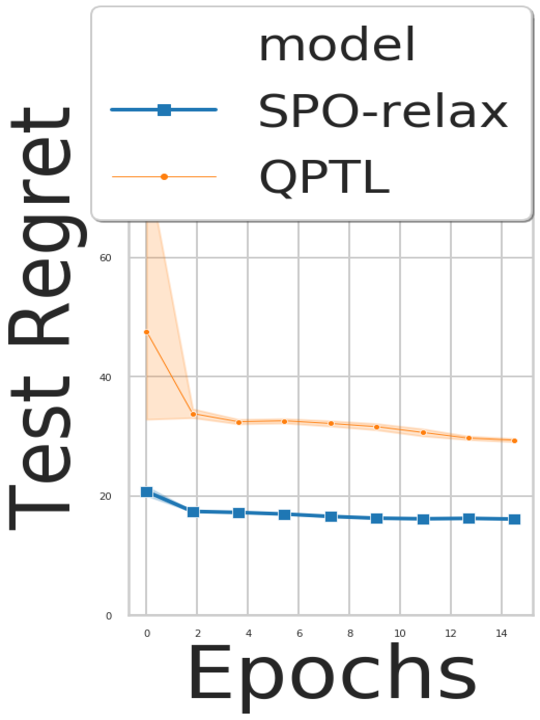}
         \caption{}
         \label{fig:unweighted_SPOvsQPT_lc_10}
     \end{subfigure} %
     ~
     \begin{subfigure}[t!]{0.14\columnwidth}
         \centering
         \includegraphics[width=\columnwidth]{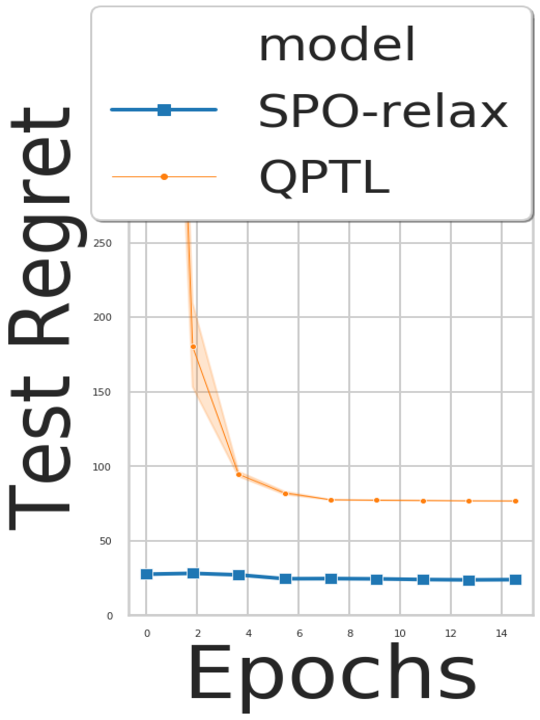}
         \caption{}
         \label{fig:unweighted_SPOvsQPTL_lc_20}
     \end{subfigure}%
     ~
     \begin{subfigure}[t!]{0.14\columnwidth}
         \centering
         \includegraphics[width=\columnwidth]{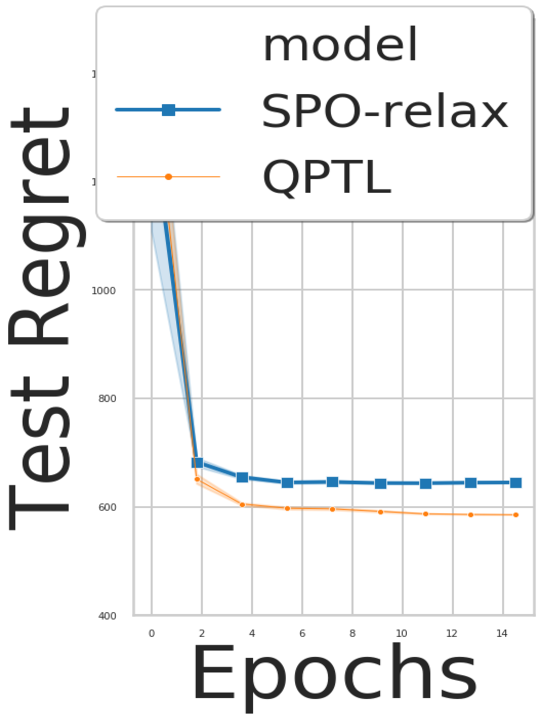}
         \caption{}
         \label{fig:weighted_SPOvsQPTL_lc_60}
     \end{subfigure} %
     ~
     \begin{subfigure}[t!]{0.14\columnwidth}
         \centering
         \includegraphics[width=\columnwidth]{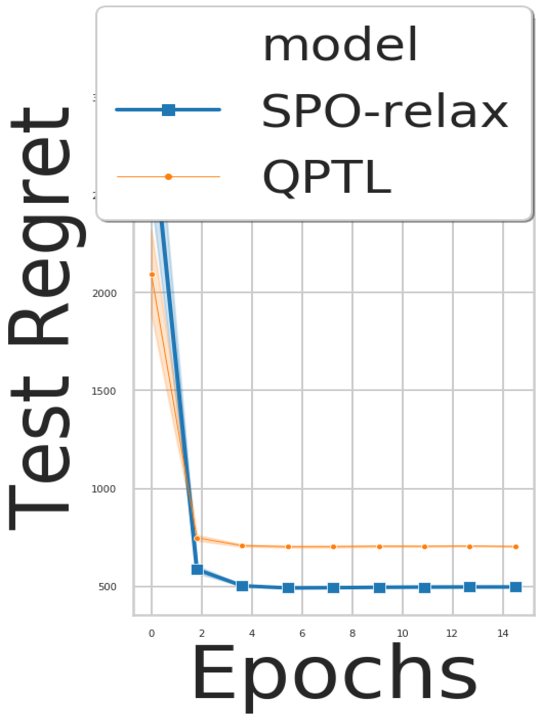}
         \caption{}
         \label{fig:weighted_SPOvsQPTL_lc_120}
     \end{subfigure}%
     ~
     \begin{subfigure}[t!]{0.14\columnwidth}
         \centering
         \includegraphics[width=\columnwidth]{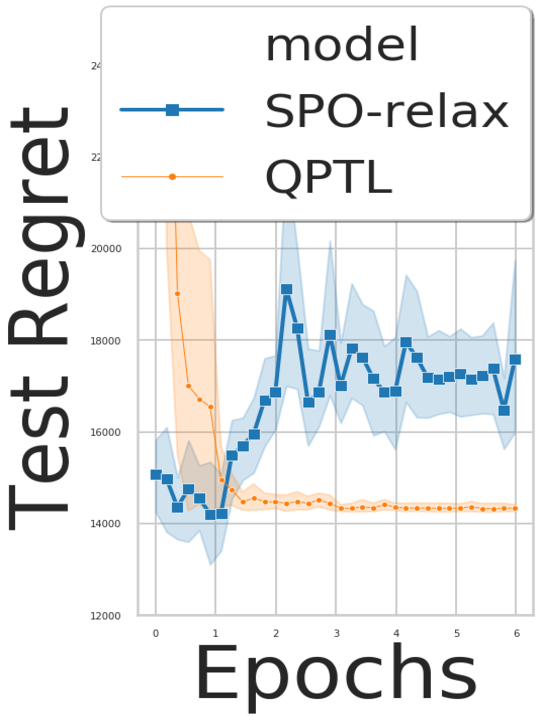}
         \caption{}
         \label{fig:energy_SPOvsQPTL_lc_l1}
     \end{subfigure}%
     ~
     \begin{subfigure}[t!]{0.14\columnwidth}
         \centering
         \includegraphics[width=\columnwidth]{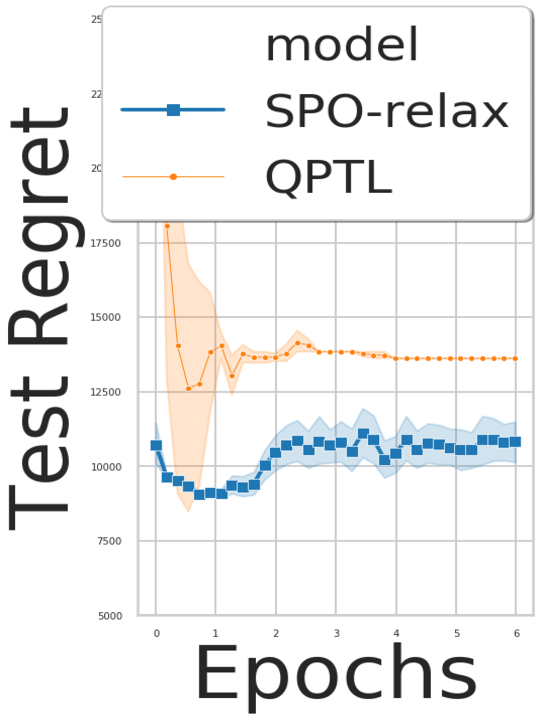}
         \caption{}
         \label{fig:energy_SPOvsQPTL_lc_l2}
     \end{subfigure}%
     \caption{Learning Curves of SPO-relax vs QPTL \\
     a: Unweighted(cap:10), b: Unweighted(cap:20), c: Weighted(cap:60), d: Weighted(cap:120), e: Energy Scheduling(I), f: Energy Scheduling(II) }
     \label{fig:SPOvsQPTL_lc}
\end{figure}

The same narrative is reflected in Figure~\ref{fig:SPOvsQPTL_lc}. In Figure~\ref{fig:weighted_SPOvsQPTL_lc_60} (weighted knapsack, capacity:60) QPTL converges to a better solution than SPO. In all other cases SPO-relax produces better quality of solution and in most cases QPTL converges slower than SPO-relax.
The poor quality of QPTL at higher capacities may stem from the squared norm which favors sparse solutions, while at high capacities, most items will be included and the best solutions are those that identify which items \textit{not} to include. 

On two energy-scheduling instances SPO-relax performs better whereas for the other instance, the regrets of SPO-relax and QPTL are similar. From Figure~\ref{fig:energy_SPOvsQPTL_lc_l2} and \ref{fig:energy_SPOvsQPTL_lc_l1}, we can see, again, SPO-relax converges faster than QPTL.

\subsection{RQ4: Suitability on large, hard optimisation instances}
\begin{table}
\resizebox{\columnwidth}{!}{
\begin{tabular}{c|rrrr|rrr}
& \multicolumn{4}{c|}{MSE-r} & \multicolumn{3}{c}{SPO-relax} \\
\cline{2-8}
Instance & {2 epochs} & {4 epochs} &{6 epochs}& {8 epochs}&{2 hour}& {4 hour} & {6 hour}  \\
\hline 
1 & 90,769 & 88,952 & 86,059 & 86,464  & 72,662 & 74,572 & 79,990   \\
2 & 128,067 & 124,450 & 124,280 & 123,738 &  120,800 & 110,944 & 114,800  \\
3 & 129,761 & 128,400 & 122,956 & 119,000 &  108,748 & 102,203 & 112,970 \\
4 & 135,398 & 132,366 & 132,167 & 126,755  & 109,694 & 99,657 & 97,351  \\
5 & 122,310 & 120,949 & 122,116 & 123,443 & 118,946 & 116,960 & 118,460 \\
\end{tabular} 
}
\caption{Relaxed regret on hard ICON challenge instances \label{tab:icon}}
\end{table}
While SPO-relax performs well across the combinatorial instances used so far, these are still toy-level problems with relatively few decision variables that can be solved in a few seconds.

We will use the large-scale optimization instances of the ICON challenge, for which no exact solutions are known. Hence, for this experiment we will report the regret when solving the relaxation of the problem for the test instances, rather than solving the discrete problem during testing as in the previous experiments.

We impose a \textit{timelimit} of $6$ hours on the total time budget that SPO-relax can spend on calling the solver. This includes the time to compute and cache the ground-truth solution of a training instance, and the timing of solving for each backpropagation of a training instance. The remaining time spent on handling the data and backpropagation is negligible in respect to the solving time.

The results are shown in Table~\ref{tab:icon}, for $5$ hard scheduling instances. First, we show the test (relaxed) regret after $2$, $4$, $6$ and $8$ MSE-r epochs. The results show that the test regret slightly decreases over the epochs; thereafter, we observed, regret tends to increase. 

With SPO-relax, in $6$ hours, it was possible to train only on $300$ to $450$ different instances, which is only $50$ to $80\%$ of the training instances. Table~\ref{tab:icon} shows even for a limited solving budget of $6$ hour and without MSE-warmstarting, it already outperforms the MSE learned models. 

This shows that even on very large and hard instances that are computationally expensive to solve, training with SPO-relax on a limited time-budget is better than training in a two-stage approach with a non optimisation-directed loss.

\section{Conclusions and future work}
Smart ``Predict and Optimize" methods have shown to be able to learn from, and improve task loss. 
Extending these techniques to be applicable beyond toy problems, more specifically hard combinatorial problems, is essential to the applicability of this promising idea.

SPO is able to outperform QPTL and lends itself to a wide applicability as it allows for the use of black-box oracles in its loss computation. We investigated the use of weaker oracles and showed that for the problems studied, learning with SPO loss while solving the relaxation leads to equal performance as solving the discrete combinatorial problem. We have shown how this opens the way to solve larger and more complex combinatorial problems, for which solving the exact solution may not be possible, let alone to do so repeatedly. 

In case of problems with weaker relaxations, one could consider adding cutting planes prior to solving~\cite{mipaal}. Moreover, further improvements could be achieved by exploiting the fact that all previously computed solutions are valid candidates. So far we have only used this for warmstarting the solver.

Our work hence encourages more research into the use of weak oracles and relaxation methods, especially those that can benefit from repeated solving. One interesting direction are local search methods and other iterative refinement methods, as they can improve the solutions during the loss computation. With respect to exact methods, knowledge compilation methods such as (relaxed) BDDs could offer both a runtime improvement from faster repeat solving and employing a relaxation.
\section{ Acknowledgments}
We would like to thank the anonymous reviewers for the valuable comments and suggestions. This research is supported by Data-driven logistics (FWO-S007318N).
\bibliographystyle{aaai}
\bibliography{citation}
\end{document}